\pdfoutput=1

\documentclass[11pt]{article}

\usepackage[]{acl}

\usepackage{times}
\usepackage{latexsym}

\usepackage{tikz}
\usepackage{comment}
\usepackage{amsmath} 
\usepackage{color}

\usepackage{xcolor}
\usepackage{url}
\usepackage{framed}
\usepackage{ragged2e}
\usepackage{bm}
\usepackage{float}
\usepackage{wrapfig}
\usepackage{bbding}
\usepackage{soul}
\usepackage{latexsym}
\usepackage{graphicx}
\usepackage{float}
\usepackage{longtable}
\usepackage{booktabs} 
\usepackage{multirow}
\usepackage{lipsum}
\usepackage{amssymb}
\usepackage{dashbox}%
\usepackage{multirow}
\usepackage{textcomp}
\usepackage{tcolorbox}
\usepackage{makecell}
\usepackage[ruled, vlined, linesnumbered]{algorithm2e}
\usepackage{mathtools}
\usepackage[ruled,vlined]{algorithm2e}
\usepackage{color, soul}

\usepackage[T1]{fontenc}

\usepackage[utf8]{inputenc}

\usepackage{microtype}

\title{PEVL: Position-enhanced Pre-training and Prompt Tuning for Vision-language Models}

\author{Yuan Yao$^{1}\thanks{\hspace{0.7em}indicates equal contribution}$\hspace{0.36em}, Qianyu Chen$^{1*}$,  Ao Zhang$^{2}$, Wei Ji$^{2}$\\ \textbf{Zhiyuan Liu$^{1\dagger}$, Tat-Seng Chua$^{2}$, Maosong Sun$^{1\dagger}$}\\
$^{1}$Dept. of Comp. Sci. \& Tech., Institute for AI, Tsinghua University, Beijing, China\\
Beijing National Research Center for Information Science and Technology\\
Innovation Center of Tsinghua University, Shanghai, China\\
$^{2}$Sea-NExT Joint Lab, Singapore\\
School of Computing, National University of Singapore, Singapore\\
\texttt{yaoyuanthu@163.com \hspace{1em} socqyc@gmail.com}
}

\begin{document}
\maketitle
\begin{abstract}
Vision-language pre-training (VLP) has shown impressive performance on a wide range of cross-modal tasks, where VLP models without reliance on object detectors are becoming the mainstream due to their superior computation efficiency and competitive performance. However, the removal of object detectors also deprives the capability of VLP models in explicit object modeling, which is essential to various position-sensitive vision-language (VL) tasks, such as referring expression comprehension and visual commonsense reasoning. To address the challenge, we introduce PEVL that enhances the pre-training and prompt tuning of VLP models with explicit object position modeling. Specifically, PEVL reformulates discretized object positions and language in a unified language modeling framework, which facilitates explicit VL alignment during pre-training, and also enables flexible prompt tuning for various downstream tasks. We show that PEVL enables state-of-the-art performance of detector-free VLP models on position-sensitive tasks such as referring expression comprehension and phrase grounding, and also improves the performance on position-insensitive tasks with grounded inputs. We make the data and code for this paper publicly available at \url{https://github.com/thunlp/PEVL}.
\end{abstract}

{\let\thefootnote\relax\footnotetext{$^\dagger$ Corresponding authors: Z.Liu (liuzy@tsinghua.edu.cn), M.Sun (sms@tsinghua.edu.cn)}}

\section{Introduction}
Recent progress on self-supervised learning has led to powerful vision-language pre-training (VLP) models that achieve state-of-the-art performance on a wide range of cross-modal tasks~\cite{lu2019vilbert,li2020oscar,radford2021learning,zhang2021vinvl,kamath2021mdetr}. Typically, VLP models are first pre-trained on large-scale image-text data to learn universal cross-modal representations, and then fine-tuned to adapt to downstream tasks~\cite{bommasani2021opportunities}. While most traditional VLP models heavily rely on external object detectors to obtain the visual inputs~\cite{lu2019vilbert,su2019vl,li2020oscar,zhang2021vinvl}, recently there is a growing interest in VLP models that remove the reliance on object detectors due to their superior computation efficiency and competitive performance~\cite{li2021align,kim2021vilt,radford2021learning,kamath2021mdetr}.

\begin{figure*}[t]
    \centering
    \includegraphics[width=\textwidth]{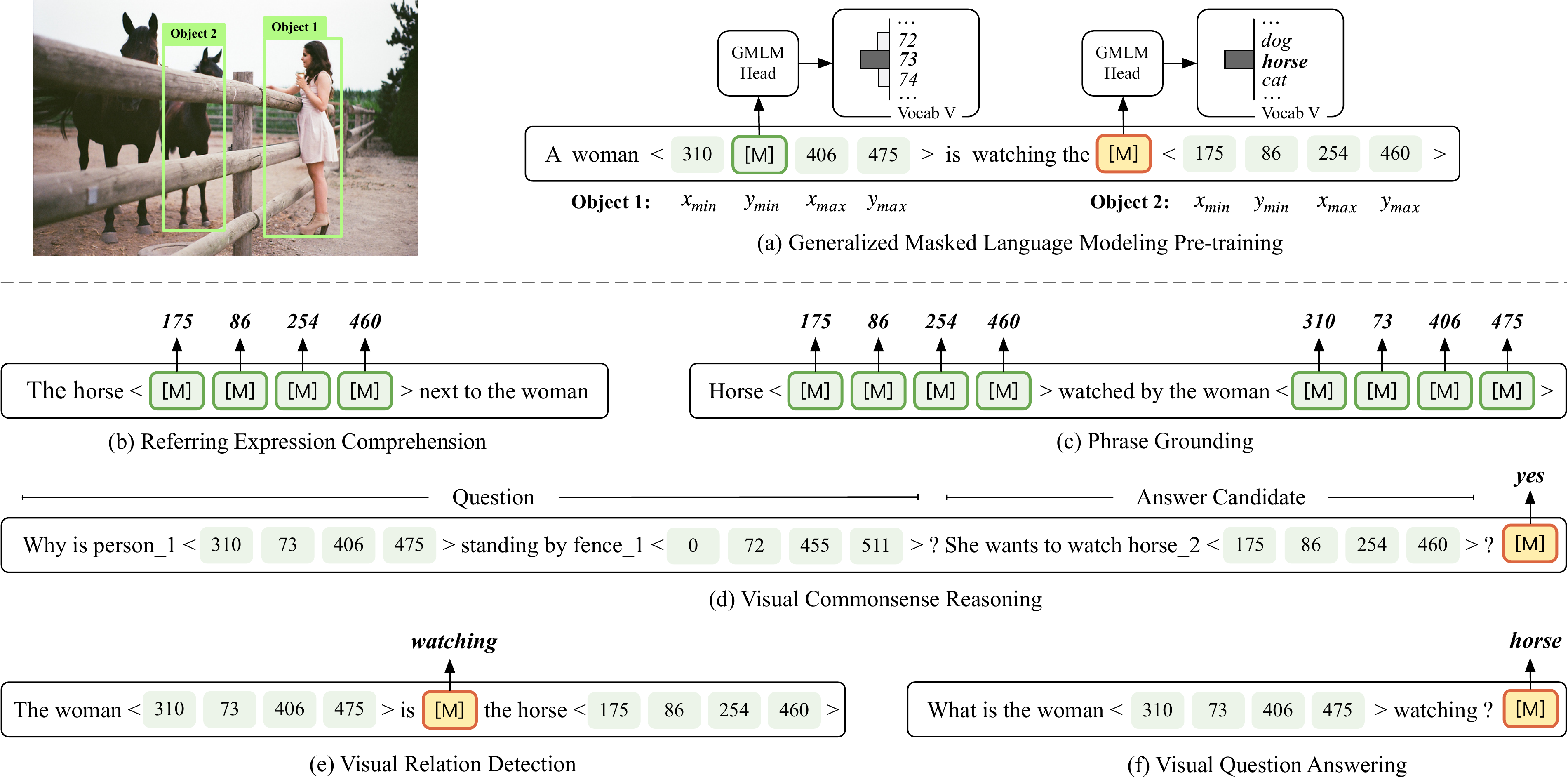}
    \caption{PEVL formulates positions and language into a unified language modeling framework. (a) During pre-training, PEVL recovers masked text and position tokens in a generalized masked language modeling (GMLM) task. (b) During prompt-tuning, PEVL reformulates various VL tasks into a fill-in-the-blank problem, which are addressed by the reused GMLM head. }
    \label{fig:framework}
\end{figure*}

However, the removal of object detectors also deprives the capability of VLP models in explicit object modeling. The drawback hinders successful handling of vision-language (VL) tasks which are inherently object-centric, where deep understanding of objects and their interactions plays an essential role~\cite{antol2015vqa,plummer2015flickr30k,krishna2017visual,hudson2019gqa}. Therefore, it is typically difficult for detector-free\footnote{Note that by detector-free, we mean that no external object detector tools are required. However, object annotations may still be needed during pre-training and tuning.} VLP models to handle various position-sensitive tasks (i.e., tasks that demand explicit object positions as input or output), such as visual commonsense reasoning~\cite{zellers2019recognition}, visual relation detection~\cite{krishna2017visual}, referring expression comprehension~\cite{yu2016modeling} and phrase grounding~\cite{plummer2015flickr30k}, which greatly undermines their generality and practicality as foundation models~\cite{bommasani2021opportunities}. For tasks that do not require explicit object modeling, such as visual question answering~\cite{antol2015vqa}, previous works have shown that introducing explicit grounding can also lead to better performance and robustness~\cite{anderson2018bottom,huang2019multi}, which can hardly be achieved in current detector-free VLP models.

In a preliminary exploration, MDETR~\cite{kamath2021mdetr} proposes to enhance detector-free VLP models by regressing object positions with Transformer decoders, serving position-output tasks such as referring expression comprehension. However, it is still unknown how to deal with various position-input tasks, such as visual commonsense reasoning and visual relation detection. Moreover, during fine-tuning, task-specific classification heads are typically introduced, resulting in a significant gap between pre-training and fine-tuning, which hinders taking full advantage of pre-trained model capabilities in downstream tasks.

In this work, we propose PEVL that enhances the pre-training and prompt tuning of VLP models with explicit object position modeling. Inspired by the recent Pix2Seq~\cite{chen2021pix2seq} that casts object detection as a language modeling task, PEVL reformulates object positions as discrete tokens, and learns the joint distribution of object positions and language in a unified language modeling framework, as shown in Figure~\ref{fig:framework}. Specifically, PEVL exploits explicit region-text alignments in existing VL datasets. Discretized position tokens are placed after object text tokens to indicate the object locations in both pre-training and prompt tuning:

(1)~During \textit{pre-training}, PEVL learns explicit VL alignment based on a generalized masked language modeling (GMLM) task, where the model recovers masked text tokens and position tokens from cross-modal context. We note that although the discretization of positions enables their unified modeling with language, it also eliminates the ordering of positions as compared with traditional continuous regression methods (i.e., predicting nearby and faraway positions to ground-truth are equally punished). The problem is exacerbated by the inevitable small disturbances in the human annotation of bounding boxes. To address the challenge, we present a novel ordering-aware objective for masked position reconstruction, which assigns larger probabilistic soft labels for nearby position tokens, and therefore retains the ordering. (2)~During \textit{prompt tuning}, PEVL can support various downstream VL tasks in a flexible prompt tuning framework, where VL tasks are addressed by the reused GMLM head in a fill-in-the-blank paradigm. In this way, PEVL maximally mitigates the gap between pre-training and tuning, and better stimulates the pre-trained model capabilities.

We conduct comprehensive experiments on five VL tasks, including position- output, input and insensitive tasks. Experimental results show that through position enhancement, PEVL enables state-of-the-art performance of detector-free VLP models on position-sensitive tasks such as referring expression comprehension and phrase grounding, and also improves the performance on position-insensitive tasks with grounded inputs. 

Our contributions are threefold: (1) We unify the modeling of positions and language in a language modeling framework, which enhances both pre-training and prompt tuning of VLP models. (2) We present a novel ordering-aware objective that retains the ordering of position tokens and avoids the influence of position annotation noise. (3) We conduct comprehensive experiments on five position-sensitive and insensitive VL tasks, which demonstrates the effectiveness of the proposed model.

\section{Preliminary}
\label{sec:preliminary}

In principle, the PEVL framework is orthogonal to VLP architectures and can be built on any VLP models to achieve position enhancement. In this work, without loss of generality, we adopt ALBEF~\cite{li2021align} as the model backbone, which is a representative detector-free VLP model that achieves state-of-the-art performance on many VL tasks. We briefly introduce the pre-training and fine-tuning procedure of ALBEF, and refer readers to the original paper for additional details.

\smallskip
\textbf{Pre-training.} The ALBEF architecture is composed of two unimodal encoders followed by a cross-modal encoder. Images and text are first encoded using a vision Transformer~\cite{dosovitskiy2020image} and a text Transformer~\cite{vaswani2017attention} respectively, and then fused with a cross-modal Transformer. The model is pre-trained with three tasks, including masked language modeling, image-text contrastive learning and image-text matching. (1) Masked language modeling aims to recover masked text tokens from the cross-modal context. (2) Image-text contrastive learning aligns the intermediate unimodal representations of image-text pairs by a contrastive loss. (3) Image-text matching classifies whether an image-text pair is aligned based on the \texttt{[CLS]} token of the cross-modal Transformer. To alleviate the noise in the pre-training text, a momentum model is maintained based on the moving-average of model parameters to provide pseudo-targets as additional supervision.

\smallskip
\textbf{Fine-tuning.} During fine-tuning, ALBEF introduces new classification heads or decoders to handle VL tasks, which leads to significant gap from pre-training. The gap hinders taking full advantage of pre-trained capabilities for downstream tasks. Moreover, since object positions cannot be explicitly modeled, detector-free VLP models typically struggle on position-sensitive tasks, which greatly undermines their generality and practicality.

\section{Methodology}
We introduce the PEVL framework, including the position reformulation for VL models, and position-enhanced VL pre-training and prompt tuning.

\subsection{Reformulating Positions for VL Models}
\label{sec:position reformulation}
Cross-modal position modeling that explicitly connects image regions and text units underpins a broad range of VL tasks. To enable strong cross-modal position modeling capability of VLP models, a primary challenge is to find a good position formulation that can be (1) easily integrated and unified into mainstream VLP models, and can be (2) flexibly prompt-tuned in various downstream tasks with minimal gap from pre-training as well. 

To this end, previous works attempt to indicate image regions by introducing region embeddings~\cite{DBLP:conf/icml/ChoLTB21} or colors~\cite{yao2021cpt} that correspond to pre-defined text tokens, which require pre-detected image regions from costly external object detectors. In contrast, we note that for the mainstream VLP models with vision Transformers ~\cite{dosovitskiy2020image} as visual encoders, image patch positions are already well indicated by positional embeddings~\cite{vaswani2017attention}, and therefore no special treatments are in fact needed for visual position coordination. 

To explicitly express visual positions in text, inspired by Pix2Seq~\cite{chen2021pix2seq} that casts object detection as a language modeling task, PEVL reformulates object bounding box coordinates as \textit{discrete position tokens}. The position tokens can be easily unified with text tokens in a language modeling framework, where the vocabulary includes both text and position tokens, and can also be easily pre-trained with existing VLP techniques. In addition to the convenience in pre-training, another important advantage is that VLP models can be easily prompt-tuned to handle various position- sensitive and insensitive VL tasks with minimal gap from pre-training, as shown in Figure~\ref{fig:framework}.

Specifically, given an image-text pair for pre-training $(I, T)$, we exploit the composing object texts and their bounding boxes $O=\{(c_i, b_i)\}_{i=1}^{N}$, where $c_i$ is the object text (e.g., \textit{person}) in text $T$, and $b_i = (x_\text{min}, y_\text{min}, x_\text{max}, y_\text{max})$ is the coordinates of the corresponding bounding box. The bounding box coordinates are discretized into position tokens as $\lfloor Mx/w \rfloor$ and $\lfloor My/h \rfloor$, where $w$ and $h$ are the width and height of the image, and $M$ is the 
total number of the position tokens. Intuitively, a larger number of position tokens will lead to a coordinate system with higher resolution, but will be more compute- and data- expensive to learn. Finally the position tokens are placed after the corresponding object text $c_i$ in $T$ to explicitly indicate the object position. Note that two special tokens ``\texttt{<}'' and ``\texttt{>}'' are introduced to indicate the start and end of position tokens, which are useful in prompting models to produce position tokens in position-output tasks. 


\subsection{Position-enhanced VL Pre-training}

After unifying positions and text in a language modeling framework, PEVL can be easily integrated into existing VLP models. To effectively learn the position and text interactions, in addition to the image-text contrastive and image-text matching tasks (see Section~\ref{sec:preliminary}), we present a novel generalized masked language modeling (GMLM) pre-training task, which recovers both masked text and position tokens based on a generalized vocabulary $\mathcal{V}$ that includes both types of tokens. We introduce two main components of the GMLM task, including masking strategy and reconstruction objective.


\smallskip

\textbf{Masking Strategy.} While the text tokens are usually masked with low ratios (e.g., 15\%) in traditional MLM tasks~\cite{devlin2019bert,li2021align}, we find that the same masking strategy cannot well serve position modeling. The reason is that object positions are relatively low-level signals, and therefore models can easily reconstruct the masked position tokens when the masking ratio is low. For example, reconstructing a single masked position token (e.g., $x_\text{min}$) given the other three unmasked ones will be largely equivalent to enclosing an object by moving a corner of the bounding box in a straight line, which does not require deep understanding of the VL semantics. Similar problems are also discussed in self-supervised learning on images~\cite{he2021masked}. 

To address the issue, we adopt \textit{high masking ratios} for position tokens, and encourage masking a more complete subset of object positions. Specifically, for each object, we randomly mask $n$ of its four position tokens with $0.25$ probability, where $n=1,2,3,4$. For example, for 25\% of the time, the four object positions  (i.e., $n=4$) are completely masked for reconstruction. For text tokens, we follow the $15\%$ masking strategy in previous works~\cite{li2021align}.\footnote{For the chosen text tokens, the replacements are 80\% \texttt{[MASK]} tokens, 10\% random tokens, and 10\% unchanged.} In this way, models are forced to learn high-level semantic interactions among image regions, text and position tokens.

\smallskip
\textbf{Reconstruction Objective.} Traditional MLM tasks typically adopt a one-hot target for token reconstruction. However, we note that the one-hot target essentially eliminates the ordering of the positions: If the position prediction is not exactly correct, predicting nearby and faraway positions to ground-truth are equally punished. The problem is exacerbated by the inevitable small disturbances in the human annotation process of bounding boxes, which confuses models in discrete position learning. To address the problem, we present a novel \textit{ordering-aware objective} for position reconstruction that assigns larger probabilistic soft labels for nearby position tokens. Specifically, given a masked position token, the unnormalized probabilistic label $y_i$ for each position token $p_i$ decreases exponentially with its distance to the ground-truth:
\begin{equation}
\small
    y_i = e^{- \alpha |p_i - p^\star|},
\end{equation}
where $|p_i - p^\star|$ is the distance between $p_i$ and the ground-truth $p^\star$, and $\alpha$ is a hyperparameter controlling the decay rate. The normalized probabilistic label $\tilde{y_i}$ is then used to compute the ordering-aware objective for position tokens:

\begin{equation}
\small
\label{eq:soft_label}
    \mathcal{L}_p = - \sum_{p_i} \tilde{y_i} \log P(\texttt{[MASK]}=p_i).
\end{equation}

The probability of position tokens is given by the GMLM head as:

\begin{equation}
\small
    P(\texttt{[MASK]}=p_i) =  \frac{\exp(\mathbf{h}_{\texttt{[MASK]}}^\top \mathbf{p}_i)}{\sum_{p_j}\exp(\mathbf{h}_{\texttt{[MASK]}}^\top \mathbf{p}_j)},
\end{equation}
where $\mathbf{h}_{\texttt{[MASK]}}$ is the hidden representation of the \texttt{[MASK]} token, and $\mathbf{p}_i$ is the representation of position token $p_i$ in the GMLM head. In this way, the objective retains the ordering of position tokens and avoids the influence of position annotation noise. For the text token reconstruction loss $\mathcal{L}_t$, we follow the traditional implementation in ALBEF~\cite{li2021align}. The final GMLM loss is the weighted sum of the loss for position token reconstruction and text token reconstruction: $\mathcal{L}_\text{GMLM} = \lambda \mathcal{L}_\text{p} + \mathcal{L}_\text{t}$, where $\lambda$ is a weighting hyperparameter.


\smallskip
\textbf{Pre-training Corpora.} PEVL exploits explicit object position annotation in VL datasets for position learning. The pre-training corpora consist of referring expressions~\cite{yu2016modeling,mao2016generation}, Flickr30k~\cite{plummer2015flickr30k}, GQA~\cite{hudson2019gqa}, VCR~\cite{zellers2019recognition} and Visual Genome~\cite{krishna2017visual}, with $4.7$M image-text pairs in total. Following~\citet{chen2020uniter}, we remove the images in the downstream test and validation sets from the pre-training corpora.

\subsection{Position-enhanced VL Prompt Tuning}
\label{sec:prompt tuning}

To adapt VLP models to downstream tasks, previous works typically introduce new classification heads or even Transformer decoders~\cite{kamath2021mdetr,li2021align,chen2020uniter}, leading to significant gap from pre-training. Recent works in pre-trained language models have shown that a consistent tuning approach with pre-training (i.e., prompt tuning) can better stimulate the pre-trained capability in downstream tasks~\cite{schick-schutze-2021-just,gao2021making,liu2021pre}. However, it is still unknown whether and how VLP models can be prompt tuned to support both position- sensitive and insensitive VL tasks.

In this context, a crucial advantage of unifying positions with language is that, VLP models can be easily prompt-tuned to handle various VL tasks based on the \textit{reused GMLM head} with minimal gap from pre-training. We divide VL tasks according to the role of positions, including position-output tasks, position-input tasks, and position-insensitive tasks. Here we introduce the main prompt tuning procedure for position-sensitive tasks. In our experiments, we show that position-insensitive tasks can also benefit from well-grounded inputs in PEVL framework (see Section~\ref{sec:main results}).

\smallskip
\textbf{Position-output Tasks} demand positions as task outputs (e.g., predicting the positions of objects described by text), such as referring expression comprehension and phrase grounding. To handle position-output tasks, we simply place four consecutive \texttt{[MASK]} tokens wrapped by ``\texttt{<}'' and ``\texttt{>}'' after object texts to be grounded for position prediction. (1) \textit{Referring Expression Comprehension.} Since the task requires locating the head noun, we place the mask tokens after the first object text for position prediction. (2) \textit{Phrase Grounding.} Since the task requires locating all objects, mask tokens are placed after each object text. After placing mask tokens, the model is prompt-tuned to produce position tokens with reused GMLM head based on the ordering-aware objective as in Equation~\ref{eq:soft_label}.

\smallskip
\textbf{Position-input Tasks} require a mixture of position and text (i.e., grounded text) as task inputs, such as visual commonsense reasoning and visual relation detection. To handle position-input tasks, PEVL first explicitly indicates the object positions in input text (see Section~\ref{sec:position reformulation})\footnote{We omit the position tokens in the task input in the following for simplicity.}, and then produces answers in a fill-in-the-blank paradigm based on the reused GMLM head. 

\smallskip
\textit{Visual Commonsense Reasoning.} Given a question, models are asked to choose the answer sentence (and rationale) from multiple candidates. For answer selection, the question $q$ and answer candidate $a_i$ are put in a prompt template as: ``$q\hspace{1.2mm} a_i \hspace{1.0mm} \text{answer:} \texttt{[MASK]}$''. Then the model can be prompted to decide which token $t \in\{\textit{yes}, \textit{no}\}$ is more proper to reconstruct the \texttt{[MASK]} token. Another plausible alternative is to reuse the image-text matching head to discriminate whether the image is aligned with the concatenated question and answer. The intuition is that a question concatenated with the correct answer can better match the image content than concatenated with a wrong answer. In our experiments, we find that the latter approach yields better performance on VCR. Despite the essential equivalence of the two prompting approaches (i.e., classifying special tokens in the last layer into binary labels with reused pre-trained heads), image-text matching task focuses more on the holistic matching between cross-modal signals during pre-training, which better fits the VCR task containing typically long text answers.

\begin{table*}[t]
\centering
\resizebox{\linewidth}{!}{%
\small
\begin{tabular}{lc ccc ccc cc cc}
\toprule
\multirow{2}{*}{Model} & \multirow{2}{*}{\makecell[c]{Object \\ Detector}} & \multicolumn{3}{c}{RefCOCO} & \multicolumn{3}{c}{RefCOCO+} & \multicolumn{2}{c}{RefCOCOg} & \multicolumn{2}{c}{Flickr30k}\\ 
\cmidrule(lr){3-5} \cmidrule(lr){6-8} \cmidrule(lr){9-10} \cmidrule(lr){11-12} &  & val  & testA & testB & val & testA & testB & val & test & val & test \\ 
\midrule
 MAttNet~\cite{yu2018mattnet} & w/ & 76.7 & 81.1 & 70.0 & 65.3 & 71.6 & 56.0 & 66.6 & 67.3 & - & -   \\
  DDPN~\cite{yu2018rethinking} & w/ & 76.8 & 80.1 & 72.4 &  64.8 & 70.5 & 54.1 & - & - & 72.8 & 73.5   \\
 VL-T5~\cite{DBLP:conf/icml/ChoLTB21} & w/ & - & - & - & -  & -  & -  & 71.2 & 71.3 & - & -  \\
 ViLBERT~\cite{lu2019vilbert} & w/ & - & - & - & 72.3 & 78.5 & 62.6 & - & - & - & -   \\
 VL-BERT\_L~\cite{su2019vl} & w/ & - & - & - & 72.6 & 78.6 & 62.3 & - & -  & - & -  \\
 UNITER\_L~\cite{chen2020uniter} & w/ & 81.4 & 87.0 & 74.2 & 75.9 & 81.5 & 66.7 & 74.9 & 75.8 & - & -   \\
 VinVL\_L~\cite{zhang2021vinvl} & w/ & 81.8 & 87.2 & 74.3 & 74.5 & 80.8 & 64.3 & 74.6 & 75.7 & - & -   \\
 VILLA\_L~\cite{DBLP:conf/nips/Gan0LZ0020} & w/ & 82.4 & 87.5 & 74.9 & 76.2 & 81.5 & 66.8 & 76.2 & 76.7 & - & -   \\
 ERNIE-ViL\_L~\cite{yu2020ernie} & w/ & - & - & - & 80.0 & 82.1 & 66.9 & - & - & - & -   \\
 UniTAB ~\cite{yang2021crossing} & w/o & 88.6 & 91.1 & 83.8 & 81.0 & 85.4 & 71.6 & 84.6 & 84.7 & - & 79.6   \\
 MDETR~\cite{kamath2021mdetr} & w/o & 87.5 & 90.4 & 82.7 & 81.1 & 85.5 & 73.0 & 83.4 & 83.3 & 82.3 & 83.8  \\
  OFA~\cite{wang2022unifying} & w/o & 88.5 &	90.7 & 83.3 & 81.4 & 87.2 & 74.3 & 	82.3	& 82.3 & - & -\\
 \midrule
 ALBEF\dag~\cite{li2021align} & w/o & - & - & - & 58.5 & 65.9 & 46.3 & - & - & - & -   \\
 PEVL & w/o & \textbf{89.6} & \textbf{92.5} & \textbf{85.0} & \textbf{83.0} & \textbf{88.4} & \textbf{74.5} & \textbf{87.1} & \textbf{86.3} & \textbf{84.1} & \textbf{84.4}   \\
 ${\rm\Delta}$ & - & - & - & - & +\textbf{24.5} & +\textbf{22.5} & +\textbf{28.2} & - & - & - & -   \\

\bottomrule
\end{tabular}}
\caption{Experimental results on referring expression comprehension and phrase grounding. L: large size model. \dag: weakly supervised results, where only image-expression pairs are used due to the lack of explicit position modeling capability. ${\rm \Delta}$: improvements of PEVL over the ALBEF backbone.}
\label{table:visual grounding}
\end{table*}

\smallskip
\textit{Visual Relation Detection.} Given an object pair $(s,o)$ (e.g., \textit{woman}, \textit{horse}) in the image, models are required to classify their semantic relation $r$ (e.g., \textit{watching}, \textit{riding}). We design the prompt template as: ``$\text{The}\hspace{1.2mm} s \hspace{1.2mm}\text{is} \hspace{0.7mm} \texttt{[MASK]} \hspace{0.7mm} \text{the} \hspace{1.2mm} o$''. Then the model is prompted to produce the relational tokens from the relation set with reused GMLM head. To deal with relations that consist of different numbers of tokens, we pad relational tokens to a maximum length $l$, and place $l$ consecutive masks in the template for relation prediction. We also include a special relation \textit{no relation with} in the relation set, which indicates no relation between the object pair. During inference, the score of relation $r$ is given by the average log probability of non-padding tokens: $s_r = \frac{1}{|r|} \sum_{i=1}^{|r|} \log P(\texttt{[MASK]}^{(i)} = r^{(i)})$, where $\texttt{[MASK]}^{(i)}$ is the $i$-th mask token, and $r^{(i)}$ is the $i$-th token of $r$. An important advantage of prompt tuning for the task is that, the large number of long-tail relations can be better learned thanks to the rich knowledge in VLP models.




\begin{table}[t]
    \begin{center}
    \small
    \resizebox{0.999\linewidth}{!}{%
    \begin{tabular}{lc ccc ccc}
    \toprule
     Model & Q $\rightarrow$ A & QA $\rightarrow$ R & Q $\rightarrow$ AR  \\ 
     
    \midrule
    R2C &    63.8  (65.1) &  67.2   (67.3) &  43.1   (44.0) \\
    TAB-VCR &    69.9  (70.4) &  72.2  (71.7) &  50.6  (50.5) \\
    VisualBERT &  70.8  (71.6) &  73.2   (73.2) &  52.2  (52.4)\\
    ViLBERT &  72.4  (73.3) &  74.5  (74.6) &  54.0  (54.8) \\
    Unicoder-VL &  72.6  (73.4) &  74.5  (74.4) &  54.4  (54.9) \\

    B2T2 &  73.2  (74.0) &  77.1  (77.1) &  56.6  (57.1) \\
    UNITER &  74.6  (75.0) &  77.0  (77.2) &  57.8  (58.2) \\
    VL-BERT &  73.8  (75.8) &  74.4  (78.4) &  55.2  (59.7) \\
    \midrule
    ALBEF &    71.9  (72.9) &  74.5  (74.5) &  54.1  (54.7) \\    
    PEVL &    75.1  (76.0) &  76.4  (76.7) &  57.8  (58.6) \\
    ${\rm\Delta}$ & +\textbf{3.2}  (+\textbf{3.1}) &  +\textbf{1.9}  (+\textbf{2.2}) &  +\textbf{3.7}  (+\textbf{3.9}) \\

    \bottomrule
    \end{tabular}}
    \caption{Experimental results of visual commonsense reasoning on VCR validation (and test) sets. }
    \label{tab:vcr}
    \end{center}
\end{table}

\section{Experiments}
We evaluate PEVL on five popular VL tasks. The models are in base size unless otherwise specified.

\subsection{Main Results}
\label{sec:main results}

\smallskip
\textbf{Referring Expression Comprehension.}
We adopt three popular datasets for the task, including RefCOCO, RefCOCO+~\cite{yu2016modeling} and RefCOCOg~\cite{mao2016generation}. We use accuracy@0.5 as the evaluation metric~\cite{kamath2021mdetr}. For baselines, we compare with state-of-the-art models for the task, and VLP models with large-size backbones. We report the weakly supervised results from ALBEF~\cite{li2021align}, which uses GRAD-CAM~\cite{selvaraju2017grad} heat map to rank the object candidates from external detectors.

\begin{table}[!t]
    \begin{center}
    \resizebox{\linewidth}{!}{%
    \small
    \begin{tabular}{l cc cc }
    \toprule
     Model &  R@50 & R@100 & mR@50 & mR@100  \\ 
     
    \midrule
    MSDN  & 64.6 & 66.6 & 15.9 & 17.5  \\

    VCTree  & 65.5 & 67.4 & 15.4 & 16.6 \\

    GPS-Net  & 65.2 & 67.1 & 15.2 & 16.6 \\

    Motif  & 66.0 & 67.9 & 14.6 & 15.8 \\
    VisualDS  & 64.4 & 66.4 &  16.1&  17.5 \\
    Unbiased  & 47.2 & 51.6 & 25.4 & 28.7 \\
    IETrans & 48.6 & 50.5 & 35.8 & 39.1\\
    DT2-ACBS  & 23.3 & 25.6 & 35.9 & 39.7 \\
    \midrule
    ALBEF & 57.6 & 63.5 & 12.2 & 15.4 \\
    PEVL & 64.4 & 66.3 & 21.7 & 23.5 \\
    ${\rm\Delta}$ & +\textbf{6.8} & +\textbf{2.8} & +\textbf{9.5} & +\textbf{8.1} \\
    \bottomrule
    \end{tabular}
    }
    \caption{Experimental results of visual relation detection on Visual Genome dataset.}
    \label{table:vrd}
    \end{center}
\vspace{-0.32em}
\end{table}

\begin{table*}[!t]
\centering

\small
\begin{tabular}{l | cc cc |cc c }
\toprule
Models & LXMERT & BAN & CTI  & CFR & ALBEF  & PEVL\dag & ${\rm\Delta}$ \\ 
\midrule
Accuracy & 59.8 & 61.5 &  61.7  & 73.6  & 64.8 & \textbf{77.0} & +\textbf{12.2} \\
\bottomrule
\end{tabular}
\caption{Visual question answering results on GQA validation set. \dag: grounded inputs.}
\label{table:vqa}
\end{table*}

\begin{table*}[!t]
\centering

\small
\begin{tabular}{l | cc cc c | c }
\toprule
Question Type & Relation & Attribute & Object & Category & Global & Overall \\ 
\midrule
Percentage (\%) & 46.7 & 32.0 & 11.8 & \hspace{1.2mm}6.5 & \hspace{1.2mm}3.1 & 100.0\\
\midrule
ALBEF & 56.9  & 67.9  & 87.9  & 62.5  &  68.1 & 64.8 \\ 
PEVL\dag & 68.4  & 84.2  & 98.1  & 68.8 & 68.5  & 77.0 \\ 
${\rm\Delta}$ & +\textbf{11.5}  & +\textbf{16.3}  & +\textbf{10.2}  & +\textbf{6.3}  & +\textbf{0.4}  & +\textbf{12.2} \\ 
\bottomrule
\end{tabular}
\caption{Visual question answering results of different question types on GQA validation set. \dag: grounded inputs.}
\label{table:question types}
\end{table*}

\begin{table*}[!t]
\centering
\resizebox{\linewidth}{!}{%
\small
\begin{tabular}{l ccc cc ccc cc c}
\toprule
\multirow{2}{*}{Model} & \multicolumn{3}{c}{RefCOCO+} & \multicolumn{2}{c}{Flickr30k} & \multicolumn{3}{c}{VCR} & \multicolumn{2}{c}{VG} & GQA \\ 
\cmidrule(lr){2-4} \cmidrule(lr){5-6} \cmidrule(lr){7-9} \cmidrule(lr){10-11} \cmidrule(lr){12-12} & val & testA & testB  & val & test & Q $\rightarrow$ A & QA $\rightarrow$ R & Q $\rightarrow$ AR & R@50 & mR@50 & val \\ 
\midrule
 PEVL & \textbf{83.1} & \textbf{88.4} & \textbf{74.5} & \textbf{84.1} & \textbf{84.4}  & 75.1 & \textbf{76.4} & \textbf{57.8} & \textbf{64.4} & \textbf{21.7} & \textbf{77.0}  \\
 \hspace{2mm} w/o PT & - & - & -  & - & - 
 & 75.1 & 76.2 & 57.6 & 61.7 & 14.2 & 77.0 \\
 \hspace{2mm} w/o OAO & 79.9 & 86.3 & 69.4  & 82.2 & 82.9  & \textbf{75.6} & 76.3 & 57.8 & 64.1 & 21.5 & 76.4 \\
 \hspace{2mm} w/o Pos & - & - & - & -  & - & 71.9 & 74.5 & 54.1 & 61.5 & 18.9 &  66.6  \\

\bottomrule
\end{tabular}}
\caption{Ablation results. PT: prompt tuning, OAO: ordering-aware objective, Pos: position tokens.}
\label{table:ablation}
\end{table*}

From the experimental results in Table~\ref{table:visual grounding}, we have the following observations: (1) PEVL outperforms all baseline models, achieving a new state-of-the-art on all three datasets for the task. Specifically, the base-size PEVL outperforms the state-of-the-art regression-based MDETR by 2.9 absolute points on the RefCOCO+ testA set, and large-size VLP models that use external detector feature inputs, such as ERNIE-ViL and VILLA. (2) PEVL significantly improves the ALBEF backbone by explicit object position modeling, effectively addressing the shortcoming in position-output tasks.


\textbf{Phrase Grounding.}
We perform experiments on the Flickr30k entities dataset~\cite{plummer2015flickr30k}. Following MDETR, we adopt merged-box accuracy@0.5 as the evaluation metric, and compare our model with the state-of-the-art baselines for the task~\cite{kamath2021mdetr,yang2021crossing}. From Table~\ref{table:visual grounding} we observe that PEVL achieves a new state-of-the-art on the phrase grounding task in grounding multiple objects in text. The results show that PEVL can effectively integrate positions with language to achieve competitive performance for various position-output tasks.

\smallskip
\textbf{Visual Commonsense Reasoning.}
We adopt the popular VCR benchmark~\cite{zellers2019recognition}, which provides human-annotated positions for objects. We report the accuracy of predicting the answer (Q $\rightarrow$ A), rationale (QA $\rightarrow$ R) and both (Q $\rightarrow$ AR). We compare with task-specific baselines and strong VLP models. For fair comparisons, we further pre-train ALBEF baseline on the same corpora as PEVL in all experiments. From the results in Table~\ref{tab:vcr}, we observe that PEVL significantly improves the ALBEF backbone (e.g., by 3.9 absolute points in Q $\rightarrow$ AR), achieving comparable performance to strong UNITER equipped with external object detectors. While the results are not state-of-the-art on the VCR benchmark, they are quite reasonable considering the current literature. The results show that PEVL can effectively provide clues for complex reasoning through grounded inputs.

\begin{figure*}[!t]
    \centering
    \includegraphics[width=\textwidth]{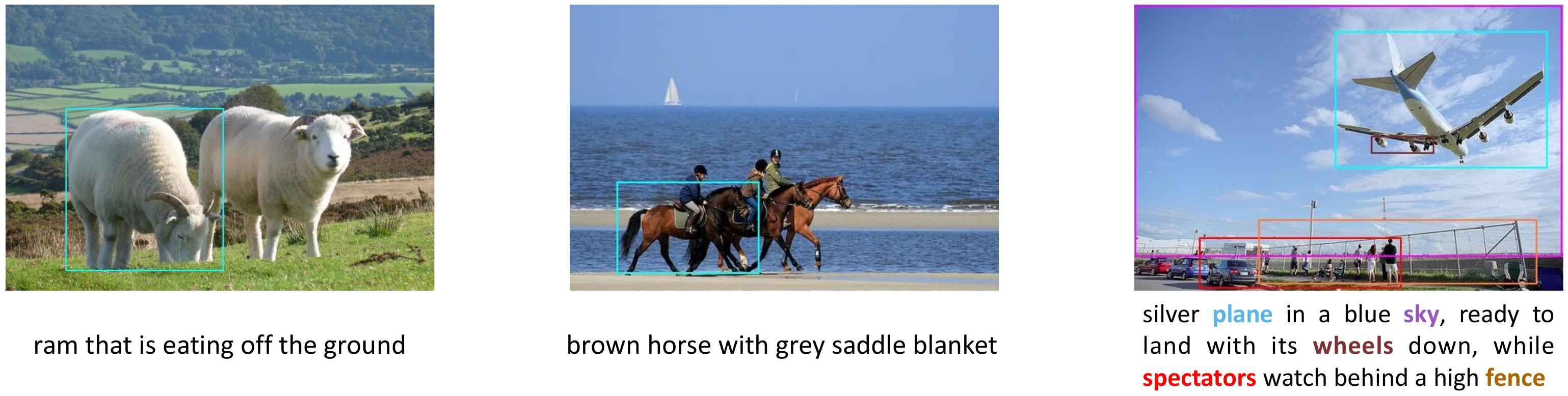}
    \caption{Case study on referring expression comprehension and phrase grounding tasks. }
    \label{fig:case}
\end{figure*}

\smallskip
\textbf{Visual Relation Detection.} We evaluate PEVL on the widely used Visual Genome dataset~\cite{krishna2017visual}, which contains 50 visual relation types. Given the human-annotated positions and labels of an object pair, models are required to predict the relations. Following previous works~\cite{tang2020unbiased,lin2020gps}, we report the recall@K (R@K) and mean recall@K (mR@K) as evaluation metrics. We compare PEVL with state-of-the-art baselines with detector feature inputs. From Table~\ref{table:vrd}, we observe that: (1) Without task-specific designs or heuristics, PEVL achieves competitive performance in both R@K and mR@K. The results show that PEVL can effectively stimulate the knowledge in VLP models for both frequent and long-tail relations through prompt tuning. (2) ALBEF struggles on the visual relation detection task, since the positions of the target object pair cannot be informed. In contrast, PEVL can effectively integrate the object position information through simple position tokens for relation prediction.

\smallskip
\textbf{Visual Question Answering.}
For position-insensitive tasks such as visual question answering, object positions are not required to be \textit{explicitly} modeled. However, we argue that explicit object position modeling can provide fine-grained clues for complex question reasoning. Specifically, we are interested in the question: Can VLP models benefit from grounded text for answering complex questions in PEVL framework?

In principle, to achieve explicit position augmentation, VQA can be decomposed into two position-sensitive stages, including an object grounding (position-output) stage and a question answering (position-input) stage. However, in our experiments, we find that the unneglectable errors in current visual grounding models constitute a bottleneck for such a two-stage model. Therefore, we turn to an ideal experiment, where the ground-truth object positions are available. Specifically, we adopt the object position annotation provided by the GQA dataset~\cite{hudson2019gqa}. We explicitly indicate the position of each object in a similar approach as in position-input tasks (see Section~\ref{sec:prompt tuning}). Then the position-enhanced question is put into the prompt template: ``$q$\hspace{0.2mm} answer: \texttt{[MASK]}''. Finally models are asked to generate answer tokens from answer candidate set. We use the same approach as in visual relation detection to cope with multi-token answers.

We report the experimental results in Table~\ref{table:vqa}, from which we observe that with grounded inputs, PEVL significantly improves the performance of ALBEF backbone in compositional question answering. The results show that object grounding is still one of the key obstacles in VQA, and PEVL can effectively utilize grounded questions for VQA in a simple prompt tuning framework. To investigate which type of questions benefit from grounded inputs, we divide the GQA validation set according to the question types from MDETR~\cite{kamath2021mdetr}. From the results in Table~\ref{table:question types}, we can see that high-quality grounding signals improve the performance on all question types. Interestingly, relation and attribute-based questions benefit more from grounded text than object-based questions, indicating the fundamental role of object modeling in reasoning over complex questions.

\subsection{Experimental Analysis}
\label{sec:analysis}

\smallskip
\textbf{Ablation Study.} We ablate key components of PEVL, including prompt tuning, ordering-aware objective, and position tokens to investigate their contribution. From the results in Table~\ref{table:ablation}, we can see that all components contribute to the final performance. Specifically, position enhancement and prompt tuning are essential for PEVL to perform position-output tasks. Prompt tuning can also be helpful in learning long-tail relations for visual relation detection, which is consistent with the results from previous works~\cite{yao2021cpt}. The ordering-aware objective contributes more to position-output tasks than position-input tasks.



\begin{table}[!t]
\centering

\small
\begin{tabular}{l | cc c |c}
\toprule
Mask & val & testA & testB  & Epochs  \\ 
\midrule
20\% & 89.3 & 92.3 &  84.3  & 5 \\
40\% & 89.0 & 92.2 &  84.7  & 5 \\
60\% & 89.4 & 92.4 &  83.4  & 7 \\
Ours & \textbf{89.6} & \textbf{92.5} &  \textbf{85.0}  & \textbf{3} \\
\bottomrule
\end{tabular}
\caption{Performance and epochs of different masking strategies on the RefCOCO dataset.}
\label{tab:mask ratio}
\end{table}

\smallskip
\textbf{Influence of Masking Strategies.} We investigate the influence of different masking strategies for position tokens during pre-training. Specifically, for baselines, the position tokens are independently chosen with a certain probability during pre-training, where the ratios of masked, replaced and unchanged tokens for the chosen token are kept identical to BERT~\cite{devlin2019bert}. We report the performance and the number of epochs required in the intermediate pre-training on the RefCOCO dataset. From the results in Table~\ref{tab:mask ratio}, we observe that our masking strategy achieves both better performance and faster convergence. The results show that a high masking ratio and a more complete subset of masked positions are both important for good position learning results.

\smallskip
\textbf{Case Study.} We visualize the position predictions on the validation sets of RefCOCO+, RefCOCOg and Flicker30k. Previous visual localization models are either based on continuous regression~\cite{kamath2021mdetr}, or limited to non-language tasks~\cite{chen2021pix2seq}. From Figure~\ref{fig:case}, we can see that discretized positions can be closely integrated with language in Transformers to achieve strong visual reasoning and localization results. Similar to regression-based models, the localization of small objects (e.g., \textit{wheels} in the right figure) can also be challenging.

\section{Related Work}


\smallskip
\textbf{Position Enhancement for VLP.} Object position modeling underpins a wide range of VL tasks. To deal with position-output tasks, some works~\cite{kamath2021mdetr,DBLP:journals/corr/abs-2104-00743,yang2021crossing} propose to perform object position prediction using Transformer decoders, but are unable to handle various position-input tasks. To explicitly indicate position inputs for VLP models, previous works explored learning region embeddings~\cite{DBLP:conf/icml/ChoLTB21}, or color-based cross-modal coreferential markers~\cite{yao2021cpt}, but rely on external object detectors. MERLOT~\cite{zellers2021merlot} also proposes to highlight objects in images with colors for position-input tasks. X-VLM~\cite{zeng2021multi} aligns multi-grained concepts in text and image for VLP models. In comparison, PEVL supports both position- input and output VL tasks in a unified prompt tuning framework.

\smallskip
\textbf{Prompt Tuning.} Prompt tuning for pre-trained language models is in rapid growth in natural language processing~\cite{petroni2019language,raffel2019exploring,brown2020language,schick-schutze-2021-just,gao2021making,qin2021learning,liu2021pre,yao2022prompt}. Recently there is also growing interest in prompt tuning VLP models. Most existing works prompt tune contrastively pre-trained image-text matching models~\cite{radford2021learning,DBLP:conf/icml/JiaYXCPPLSLD21} for recognition tasks~\cite{zhou2021learning,rao2021denseclip,wang2021actionclip,gu2021zero,xie2021zsd,ju2021prompting}. CPT~\cite{yao2021cpt} prompt tunes VLP models with color-based prompts, and achieves promising results for zero- and few-shot tasks. Some works perform pre-training and VL tasks using identical Transformer decoders in an auto-regressive fashion~\cite{DBLP:journals/corr/abs-2108-10904,DBLP:conf/icml/ChoLTB21,yang2021crossing,tsimpoukelli2021multimodal}, which avoids the gap between pre-training and tuning, but are typically limited in performance due to the unidirectional architecture.



\section{Conclusion and Future Work}
In this work, we present PEVL that enhances the pre-training and prompt-tuning of detector-free VLP models with unified position and language modeling. Comprehensive experimental results demonstrate the effectiveness of the proposed model. Future works include exploring weakly supervised signals for position and language learning without human annotation.

\section{Acknowledgement}
This work is supported by the National Key R\&D Program of China (No. 2020AAA0106502), Institute Guo Qiang at Tsinghua University and NExT++ project from the National Research Foundation, Prime Minister’s Office, Singapore under its IRC@Singapore Funding Initiative.

Yuan Yao designed the framework and experiments, and wrote
the paper. Qianyu Chen conducted the
experiments. Ao Zhang and Wei Ji participated in the discussion. Zhiyuan Liu, Tat-Seng Chua and Maosong Sun advised the project.

\section{Limitations}
We identify several key limitations of PEVL that are promising for future explorations.

\smallskip
\noindent
\textbf{Computation Cost.} To explicitly model object positions and text in a unified framework, we introduce four additional position tokens for each target object, which requires more computation cost. Improving the computation efficiency of additional position tokens is an important direction for future improvements.

\smallskip
\noindent
\textbf{Object Annotation.} Similar to other explicit object position modeling VLP models, PEVL requires manual object annotation in multi-modal datasets. It will be promising to explore pre-training tasks that can learn position tokens with less supervision. For example, VLP models can be bootstrapped from small-scale human annotations and large-scale predicted annotations.

\bibliography{custom}

\begin{thebibliography}{69}
\expandafter\ifx\csname natexlab\endcsname\relax\def\natexlab#1{#1}\fi

\bibitem[{Alberti et~al.(2019)Alberti, Ling, Collins, and
  Reitter}]{alberti2019fusion}
Chris Alberti, Jeffrey Ling, Michael Collins, and David Reitter. 2019.
\newblock \href {https://arxiv.org/abs/1908.05054} {Fusion of detected objects
  in text for visual question answering}.
\newblock In \emph{Proceedings of EMNLP-IJCNLP}.

\bibitem[{Anderson et~al.(2018)Anderson, He, Buehler, Teney, Johnson, Gould,
  and Zhang}]{anderson2018bottom}
Peter Anderson, Xiaodong He, Chris Buehler, Damien Teney, Mark Johnson, Stephen
  Gould, and Lei Zhang. 2018.
\newblock \href
  {https://openaccess.thecvf.com/content_cvpr_2018/papers/Anderson_Bottom-Up_and_Top-Down_CVPR_2018_paper.pdf}
  {Bottom-up and top-down attention for image captioning and visual question
  answering}.
\newblock In \emph{Proceedings of CVPR}.

\bibitem[{Antol et~al.(2015)Antol, Agrawal, Lu, Mitchell, Batra, Zitnick, and
  Parikh}]{antol2015vqa}
Stanislaw Antol, Aishwarya Agrawal, Jiasen Lu, Margaret Mitchell, Dhruv Batra,
  C~Lawrence Zitnick, and Devi Parikh. 2015.
\newblock \href {http://arxiv.org/pdf/1505.00468.pdf} {{VQA}: Visual question
  answering}.
\newblock In \emph{Proceedings of ICCV}.

\bibitem[{Bommasani et~al.(2021)Bommasani, Hudson, Adeli, Altman, Arora, von
  Arx, Bernstein, Bohg, Bosselut, Brunskill
  et~al.}]{bommasani2021opportunities}
Rishi Bommasani, Drew~A Hudson, Ehsan Adeli, Russ Altman, Simran Arora, Sydney
  von Arx, Michael~S Bernstein, Jeannette Bohg, Antoine Bosselut, Emma
  Brunskill, et~al. 2021.
\newblock \href {https://arxiv.org/abs/2108.07258} {On the opportunities and
  risks of foundation models}.
\newblock \emph{arXiv preprint arXiv:2108.07258}.

\bibitem[{Brown et~al.(2020)Brown, Mann, Ryder, Subbiah, Kaplan, Dhariwal,
  Neelakantan, Shyam, Sastry, Askell et~al.}]{brown2020language}
Tom~B Brown, Benjamin Mann, Nick Ryder, Melanie Subbiah, Jared Kaplan, Prafulla
  Dhariwal, Arvind Neelakantan, Pranav Shyam, Girish Sastry, Amanda Askell,
  et~al. 2020.
\newblock \href
  {https://papers.nips.cc/paper/2020/file/1457c0d6bfcb4967418bfb8ac142f64a-Paper.pdf}
  {Language models are few-shot learners}.
\newblock In \emph{Proceddings of NeurIPS}.

\bibitem[{Chen et~al.(2022)Chen, Saxena, Li, Fleet, and
  Hinton}]{chen2021pix2seq}
Ting Chen, Saurabh Saxena, Lala Li, David~J Fleet, and Geoffrey Hinton. 2022.
\newblock \href {https://openreview.net/pdf?id=e42KbIw6Wb} {{Pix2Seq}: A
  language modeling framework for object detection}.
\newblock In \emph{Proceedings of ICLR}.

\bibitem[{Chen et~al.(2020)Chen, Li, Yu, El~Kholy, Ahmed, Gan, Cheng, and
  Liu}]{chen2020uniter}
Yen-Chun Chen, Linjie Li, Licheng Yu, Ahmed El~Kholy, Faisal Ahmed, Zhe Gan,
  Yu~Cheng, and Jingjing Liu. 2020.
\newblock \href {https://arxiv.org/abs/1909.11740} {{UNITER}: Universal
  image-text representation learning}.
\newblock In \emph{Proceedings of ECCV}.

\bibitem[{Cho et~al.(2021)Cho, Lei, Tan, and Bansal}]{DBLP:conf/icml/ChoLTB21}
Jaemin Cho, Jie Lei, Hao Tan, and Mohit Bansal. 2021.
\newblock \href {http://proceedings.mlr.press/v139/cho21a/cho21a.pdf} {Unifying
  vision-and-language tasks via text generation}.
\newblock In \emph{Proceedings of ICML}.

\bibitem[{Desai et~al.(2021)Desai, Wu, Tripathi, and
  Vasconcelos}]{desai2021dt2}
Alakh Desai, Tz-Ying Wu, Subarna Tripathi, and Nuno Vasconcelos. 2021.
\newblock \href
  {https://openaccess.thecvf.com/content/ICCV2021/papers/Desai_Learning_of_Visual_Relations_The_Devil_Is_in_the_Tails_ICCV_2021_paper.pdf}
  {Learning of visual relations: The devil is in the tails}.
\newblock In \emph{Proceedings of ICCV}.

\bibitem[{Devlin et~al.(2019)Devlin, Chang, Lee, and
  Toutanova}]{devlin2019bert}
Jacob Devlin, Ming-Wei Chang, Kenton Lee, and Kristina Toutanova. 2019.
\newblock \href {https://aclanthology.org/N19-1423.pdf} {{BERT}: Pre-training
  of deep bidirectional transformers for language understanding}.
\newblock In \emph{Proceedings of NAACL-HLT}.

\bibitem[{Do et~al.(2019)Do, Do, Tran, Tjiputra, and Tran}]{do2019compact}
Tuong Do, Thanh-Toan Do, Huy Tran, Erman Tjiputra, and Quang~D Tran. 2019.
\newblock \href
  {https://openaccess.thecvf.com/content_ICCV_2019/papers/Do_Compact_Trilinear_Interaction_for_Visual_Question_Answering_ICCV_2019_paper.pdf}
  {Compact trilinear interaction for visual question answering}.
\newblock In \emph{Proceedings of ICCV}.

\bibitem[{Dosovitskiy et~al.(2021)Dosovitskiy, Beyer, Kolesnikov, Weissenborn,
  Zhai, Unterthiner, Dehghani, Minderer, Heigold, Gelly
  et~al.}]{dosovitskiy2020image}
Alexey Dosovitskiy, Lucas Beyer, Alexander Kolesnikov, Dirk Weissenborn,
  Xiaohua Zhai, Thomas Unterthiner, Mostafa Dehghani, Matthias Minderer, Georg
  Heigold, Sylvain Gelly, et~al. 2021.
\newblock \href {https://openreview.net/pdf?id=YicbFdNTTy} {An image is worth
  16x16 words: Transformers for image recognition at scale}.
\newblock In \emph{Proceedings of ICLR}.

\bibitem[{Gan et~al.(2020)Gan, Chen, Li, Zhu, Cheng, and
  Liu}]{DBLP:conf/nips/Gan0LZ0020}
Zhe Gan, Yen{-}Chun Chen, Linjie Li, Chen Zhu, Yu~Cheng, and Jingjing Liu.
  2020.
\newblock \href
  {https://proceedings.neurips.cc/paper/2020/file/49562478de4c54fafd4ec46fdb297de5-Paper.pdf}
  {Large-scale adversarial training for vision-and-language representation
  learning}.
\newblock In \emph{Proceedings of NeurIPS}.

\bibitem[{Gao et~al.(2021)Gao, Fisch, and Chen}]{gao2021making}
Tianyu Gao, Adam Fisch, and Danqi Chen. 2021.
\newblock \href {https://aclanthology.org/2021.acl-long.295.pdf} {Making
  pre-trained language models better few-shot learners}.
\newblock In \emph{Proceedings of ACL}.

\bibitem[{Gu et~al.(2022)Gu, Lin, Kuo, and Cui}]{gu2021zero}
Xiuye Gu, Tsung-Yi Lin, Weicheng Kuo, and Yin Cui. 2022.
\newblock \href {https://arxiv.org/abs/2104.13921} {Open-vocabulary object
  detection via vision and language knowledge distillation}.
\newblock In \emph{Proceedings of ICLR}.

\bibitem[{Gupta et~al.(2022)Gupta, Kamath, Kembhavi, and
  Hoiem}]{DBLP:journals/corr/abs-2104-00743}
Tanmay Gupta, Amita Kamath, Aniruddha Kembhavi, and Derek Hoiem. 2022.
\newblock \href
  {https://openaccess.thecvf.com/content/CVPR2022/papers/Gupta_Towards_General_Purpose_Vision_Systems_An_End-to-End_Task-Agnostic_Vision-Language_Architecture_CVPR_2022_paper.pdf}
  {Towards general purpose vision systems: An end-to-end task-agnostic
  vision-language architecture}.
\newblock In \emph{Proceedings of CVPR}.

\bibitem[{He et~al.(2022)He, Chen, Xie, Li, Doll{\'a}r, and
  Girshick}]{he2021masked}
Kaiming He, Xinlei Chen, Saining Xie, Yanghao Li, Piotr Doll{\'a}r, and Ross
  Girshick. 2022.
\newblock \href
  {https://openaccess.thecvf.com/content/CVPR2022/papers/He_Masked_Autoencoders_Are_Scalable_Vision_Learners_CVPR_2022_paper.pdf}
  {Masked autoencoders are scalable vision learners}.
\newblock In \emph{Proceedings of CVPR}.

\bibitem[{Huang et~al.(2019)Huang, Huang, Guo, Qiao, and Zhu}]{huang2019multi}
Pingping Huang, Jianhui Huang, Yuqing Guo, Min Qiao, and Yong Zhu. 2019.
\newblock \href {https://aclanthology.org/P19-1349.pdf} {Multi-grained
  attention with object-level grounding for visual question answering}.
\newblock In \emph{Proceedings of ACL}.

\bibitem[{Hudson and Manning(2019)}]{hudson2019gqa}
Drew~A Hudson and Christopher~D Manning. 2019.
\newblock \href
  {https://openaccess.thecvf.com/content_CVPR_2019/papers/Hudson_GQA_A_New_Dataset_for_Real-World_Visual_Reasoning_and_Compositional_CVPR_2019_paper.pdf}
  {{GQA}: A new dataset for real-world visual reasoning and compositional
  question answering}.
\newblock In \emph{Proceedings of CVPR}.

\bibitem[{Jia et~al.(2021)Jia, Yang, Xia, Chen, Parekh, Pham, Le, Sung, Li, and
  Duerig}]{DBLP:conf/icml/JiaYXCPPLSLD21}
Chao Jia, Yinfei Yang, Ye~Xia, Yi{-}Ting Chen, Zarana Parekh, Hieu Pham,
  Quoc~V. Le, Yun{-}Hsuan Sung, Zhen Li, and Tom Duerig. 2021.
\newblock \href {http://proceedings.mlr.press/v139/jia21b/jia21b.pdf} {Scaling
  up visual and vision-language representation learning with noisy text
  supervision}.
\newblock In \emph{Proceedings of ICML}.

\bibitem[{Ju et~al.(2022)Ju, Han, Zheng, Zhang, and Xie}]{ju2021prompting}
Chen Ju, Tengda Han, Kunhao Zheng, Ya~Zhang, and Weidi Xie. 2022.
\newblock \href {https://arxiv.org/pdf/2112.04478v2.pdf} {Prompting
  visual-language models for efficient video understanding}.
\newblock In \emph{Proceedings of ECCV}.

\bibitem[{Kamath et~al.(2021)Kamath, Singh, LeCun, Misra, Synnaeve, and
  Carion}]{kamath2021mdetr}
Aishwarya Kamath, Mannat Singh, Yann LeCun, Ishan Misra, Gabriel Synnaeve, and
  Nicolas Carion. 2021.
\newblock \href
  {https://openaccess.thecvf.com/content/ICCV2021/papers/Kamath_MDETR_-_Modulated_Detection_for_End-to-End_Multi-Modal_Understanding_ICCV_2021_paper.pdf}
  {{MDETR}: Modulated detection for end-to-end multi-modal understanding}.
\newblock In \emph{Proceedings of ICCV}.

\bibitem[{Kim et~al.(2018)Kim, Jun, and Zhang}]{kim2018bilinear}
Jin-Hwa Kim, Jaehyun Jun, and Byoung-Tak Zhang. 2018.
\newblock \href
  {http://papers.neurips.cc/paper/7429-bilinear-attention-networks.pdf}
  {Bilinear attention networks}.
\newblock In \emph{Proceedings of NeurIPS}.

\bibitem[{Kim et~al.(2021)Kim, Son, and Kim}]{kim2021vilt}
Wonjae Kim, Bokyung Son, and Ildoo Kim. 2021.
\newblock \href {http://proceedings.mlr.press/v139/kim21k/kim21k.pdf} {{ViLT}:
  Vision-and-language transformer without convolution or region supervision}.
\newblock In \emph{Proceedings of ICML}.

\bibitem[{Krishna et~al.(2017)Krishna, Zhu, Groth, Johnson, Hata, Kravitz,
  Chen, Kalantidis, Li, Shamma et~al.}]{krishna2017visual}
Ranjay Krishna, Yuke Zhu, Oliver Groth, Justin Johnson, Kenji Hata, Joshua
  Kravitz, Stephanie Chen, Yannis Kalantidis, Li-Jia Li, David~A Shamma, et~al.
  2017.
\newblock \href
  {https://link.springer.com/content/pdf/10.1007/s11263-016-0981-7.pdf}
  {{Visual Genome}: Connecting language and vision using crowdsourced dense
  image annotations}.
\newblock \emph{International Journal of Computer Vision}.

\bibitem[{Li et~al.(2020{\natexlab{a}})Li, Duan, Fang, Gong, and
  Jiang}]{li2020unicoder}
Gen Li, Nan Duan, Yuejian Fang, Ming Gong, and Daxin Jiang. 2020{\natexlab{a}}.
\newblock \href {https://ojs.aaai.org/index.php/AAAI/article/view/6795/6649}
  {Unicoder-{VL}: A universal encoder for vision and language by cross-modal
  pre-training}.
\newblock In \emph{Proceedings of AAAI}.

\bibitem[{Li et~al.(2021)Li, Selvaraju, Gotmare, Joty, Xiong, and
  Hoi}]{li2021align}
Junnan Li, Ramprasaath~R Selvaraju, Akhilesh~Deepak Gotmare, Shafiq Joty,
  Caiming Xiong, and Steven Hoi. 2021.
\newblock \href {https://openreview.net/pdf?id=OJLaKwiXSbx} {Align before fuse:
  Vision and language representation learning with momentum distillation}.
\newblock In \emph{Proceedings of NeurIPS}.

\bibitem[{Li et~al.(2020{\natexlab{b}})Li, Yatskar, Yin, Hsieh, and
  Chang}]{DBLP:journals/corr/abs-1908-03557}
Liunian~Harold Li, Mark Yatskar, Da~Yin, Cho{-}Jui Hsieh, and Kai{-}Wei Chang.
  2020{\natexlab{b}}.
\newblock \href {https://arxiv.org/abs/1908.03557} {{VisualBERT}: {A} simple
  and performant baseline for vision and language}.
\newblock In \emph{Proceedings of ACL}.

\bibitem[{Li et~al.(2020{\natexlab{c}})Li, Yin, Li, Zhang, Hu, Zhang, Wang, Hu,
  Dong, Wei et~al.}]{li2020oscar}
Xiujun Li, Xi~Yin, Chunyuan Li, Pengchuan Zhang, Xiaowei Hu, Lei Zhang, Lijuan
  Wang, Houdong Hu, Li~Dong, Furu Wei, et~al. 2020{\natexlab{c}}.
\newblock \href {https://arxiv.org/abs/2004.06165} {Oscar: Object-semantics
  aligned pre-training for vision-language tasks}.
\newblock In \emph{Proceedings of ECCV}.

\bibitem[{Lin et~al.(2019)Lin, Jain, and Schwing}]{LinNeurIPS2019}
J.~Lin, U.~Jain, and A.~G. Schwing. 2019.
\newblock \href
  {https://papers.nips.cc/paper/2019/file/1fa6269f58898f0e809575c9a48747ef-Paper.pdf}
  {{TAB-VCR}: Tags and attributes based vcr baselines}.
\newblock In \emph{Proceedings of NeurIPS}.

\bibitem[{Lin et~al.(2020)Lin, Ding, Zeng, and Tao}]{lin2020gps}
Xin Lin, Changxing Ding, Jinquan Zeng, and Dacheng Tao. 2020.
\newblock \href
  {https://openaccess.thecvf.com/content_CVPR_2020/papers/Lin_GPS-Net_Graph_Property_Sensing_Network_for_Scene_Graph_Generation_CVPR_2020_paper.pdf}
  {{GPS-Net}: Graph property sensing network for scene graph generation}.
\newblock In \emph{Proceedings of CVPR}.

\bibitem[{Liu et~al.(2021)Liu, Yuan, Fu, Jiang, Hayashi, and
  Neubig}]{liu2021pre}
Pengfei Liu, Weizhe Yuan, Jinlan Fu, Zhengbao Jiang, Hiroaki Hayashi, and
  Graham Neubig. 2021.
\newblock \href {https://arxiv.org/abs/2107.13586} {Pre-train, prompt, and
  predict: A systematic survey of prompting methods in natural language
  processing}.
\newblock \emph{ACM Computing Surveys}.

\bibitem[{Lu et~al.(2019)Lu, Batra, Parikh, and Lee}]{lu2019vilbert}
Jiasen Lu, Dhruv Batra, Devi Parikh, and Stefan Lee. 2019.
\newblock \href
  {https://papers.nips.cc/paper/2019/file/c74d97b01eae257e44aa9d5bade97baf-Paper.pdf}
  {{ViLBERT}: Pretraining task-agnostic visiolinguistic representations for
  vision-and-language tasks}.
\newblock In \emph{Proceedings of NeurIPS}.

\bibitem[{Mao et~al.(2016)Mao, Huang, Toshev, Camburu, Yuille, and
  Murphy}]{mao2016generation}
Junhua Mao, Jonathan Huang, Alexander Toshev, Oana Camburu, Alan~L Yuille, and
  Kevin Murphy. 2016.
\newblock \href {https://arxiv.org/abs/1511.02283} {Generation and
  comprehension of unambiguous object descriptions}.
\newblock In \emph{Proceedings of CVPR}.

\bibitem[{Nguyen et~al.(2022)Nguyen, Do, Tran, Tjiputra, Tran, and
  Nguyen}]{nguyen2021coarse}
Binh~X Nguyen, Tuong Do, Huy Tran, Erman Tjiputra, Quang~D Tran, and Anh
  Nguyen. 2022.
\newblock \href
  {https://openaccess.thecvf.com/content/CVPR2022W/MULA/papers/Nguyen_Coarse-To-Fine_Reasoning_for_Visual_Question_Answering_CVPRW_2022_paper.pdf}
  {Coarse-to-fine reasoning for visual question answering}.
\newblock In \emph{Proceedings of CVPR Workshops}.

\bibitem[{Petroni et~al.(2019)Petroni, Rockt{\"a}schel, Riedel, Lewis, Bakhtin,
  Wu, and Miller}]{petroni2019language}
Fabio Petroni, Tim Rockt{\"a}schel, Sebastian Riedel, Patrick Lewis, Anton
  Bakhtin, Yuxiang Wu, and Alexander Miller. 2019.
\newblock \href {https://aclanthology.org/D19-1250.pdf} {Language models as
  knowledge bases?}
\newblock In \emph{Proceedings of EMNLP-IJCNLP}.

\bibitem[{Plummer et~al.(2015)Plummer, Wang, Cervantes, Caicedo, Hockenmaier,
  and Lazebnik}]{plummer2015flickr30k}
Bryan~A Plummer, Liwei Wang, Chris~M Cervantes, Juan~C Caicedo, Julia
  Hockenmaier, and Svetlana Lazebnik. 2015.
\newblock \href
  {https://openaccess.thecvf.com/content_iccv_2015/papers/Plummer_Flickr30k_Entities_Collecting_ICCV_2015_paper.pdf}
  {Flickr30k entities: Collecting region-to-phrase correspondences for richer
  image-to-sentence models}.
\newblock In \emph{Proceedings of ICCV}.

\bibitem[{Qin and Eisner(2021)}]{qin2021learning}
Guanghui Qin and Jason Eisner. 2021.
\newblock \href {https://aclanthology.org/2021.naacl-main.410.pdf} {Learning
  how to ask: Querying {LM}s with mixtures of soft prompts}.
\newblock In \emph{Proceedings of NAACL}.

\bibitem[{Radford et~al.(2021)Radford, Kim, Hallacy, Ramesh, Goh, Agarwal,
  Sastry, Askell, Mishkin, Clark et~al.}]{radford2021learning}
Alec Radford, Jong~Wook Kim, Chris Hallacy, Aditya Ramesh, Gabriel Goh,
  Sandhini Agarwal, Girish Sastry, Amanda Askell, Pamela Mishkin, Jack Clark,
  et~al. 2021.
\newblock \href {http://proceedings.mlr.press/v139/radford21a/radford21a.pdf}
  {Learning transferable visual models from natural language supervision}.
\newblock In \emph{Proceedings of ICML}.

\bibitem[{Raffel et~al.(2020)Raffel, Shazeer, Roberts, Lee, Narang, Matena,
  Zhou, Li, and Liu}]{raffel2019exploring}
Colin Raffel, Noam Shazeer, Adam Roberts, Katherine Lee, Sharan Narang, Michael
  Matena, Yanqi Zhou, Wei Li, and Peter~J Liu. 2020.
\newblock \href {http://jmlr.org/papers/v21/20-074.html} {Exploring the limits
  of transfer learning with a unified text-to-text transformer}.
\newblock \emph{Journal of Machine Learning Research}.

\bibitem[{Rao et~al.(2022)Rao, Zhao, Chen, Tang, Zhu, Huang, Zhou, and
  Lu}]{rao2021denseclip}
Yongming Rao, Wenliang Zhao, Guangyi Chen, Yansong Tang, Zheng Zhu, Guan Huang,
  Jie Zhou, and Jiwen Lu. 2022.
\newblock \href {https://arxiv.org/abs/2112.01518} {{DenseCLIP}:
  Language-guided dense prediction with context-aware prompting}.
\newblock In \emph{Proceedings of CVPR}.

\bibitem[{Schick and Sch{\"u}tze(2021)}]{schick-schutze-2021-just}
Timo Schick and Hinrich Sch{\"u}tze. 2021.
\newblock \href {https://aclanthology.org/2021.naacl-main.185.pdf} {It{'}s not
  just size that matters: Small language models are also few-shot learners}.
\newblock In \emph{Proceedings of NAACL}.

\bibitem[{Selvaraju et~al.(2017)Selvaraju, Cogswell, Das, Vedantam, Parikh, and
  Batra}]{selvaraju2017grad}
Ramprasaath~R Selvaraju, Michael Cogswell, Abhishek Das, Ramakrishna Vedantam,
  Devi Parikh, and Dhruv Batra. 2017.
\newblock \href
  {https://openaccess.thecvf.com/content_ICCV_2017/papers/Selvaraju_Grad-CAM_Visual_Explanations_ICCV_2017_paper.pdf}
  {{GRAD-CAM}: Visual explanations from deep networks via gradient-based
  localization}.
\newblock In \emph{Proceedings of ICCV}.

\bibitem[{Su et~al.(2020)Su, Zhu, Cao, Li, Lu, Wei, and Dai}]{su2019vl}
Weijie Su, Xizhou Zhu, Yue Cao, Bin Li, Lewei Lu, Furu Wei, and Jifeng Dai.
  2020.
\newblock \href
  {https://openreview.net/attachment?id=SygXPaEYvH&name=original_pdf}
  {{VL-BERT}: Pre-training of generic visual-linguistic representations}.
\newblock In \emph{Proceedings of ICLR}.

\bibitem[{Tan and Bansal(2019)}]{tan2019lxmert}
Hao Tan and Mohit Bansal. 2019.
\newblock \href {https://aclanthology.org/D19-1514.pdf} {{LXMERT}: Learning
  cross-modality encoder representations from transformers}.
\newblock In \emph{Proceedings of EMNLP-IJCNLP}.

\bibitem[{Tang et~al.(2020)Tang, Niu, Huang, Shi, and Zhang}]{tang2020unbiased}
Kaihua Tang, Yulei Niu, Jianqiang Huang, Jiaxin Shi, and Hanwang Zhang. 2020.
\newblock \href
  {https://openaccess.thecvf.com/content_CVPR_2020/papers/Tang_Unbiased_Scene_Graph_Generation_From_Biased_Training_CVPR_2020_paper.pdf}
  {Unbiased scene graph generation from biased training}.
\newblock In \emph{Proceedings of CVPR}.

\bibitem[{Tang et~al.(2019)Tang, Zhang, Wu, Luo, and Liu}]{tang2019learning}
Kaihua Tang, Hanwang Zhang, Baoyuan Wu, Wenhan Luo, and Wei Liu. 2019.
\newblock \href
  {https://openaccess.thecvf.com/content_CVPR_2019/papers/Tang_Learning_to_Compose_Dynamic_Tree_Structures_for_Visual_Contexts_CVPR_2019_paper.pdf}
  {Learning to compose dynamic tree structures for visual contexts}.
\newblock In \emph{Proceedings of CVPR}.

\bibitem[{Tsimpoukelli et~al.(2021)Tsimpoukelli, Menick, Cabi, Eslami, Vinyals,
  and Hill}]{tsimpoukelli2021multimodal}
Maria Tsimpoukelli, Jacob Menick, Serkan Cabi, SM~Eslami, Oriol Vinyals, and
  Felix Hill. 2021.
\newblock \href {https://openreview.net/pdf?id=WtmMyno9Tq2} {Multimodal
  few-shot learning with frozen language models}.
\newblock In \emph{Proceedings of NeurIPS}.

\bibitem[{Vaswani et~al.(2017)Vaswani, Shazeer, Parmar, Uszkoreit, Jones,
  Gomez, Kaiser, and Polosukhin}]{vaswani2017attention}
Ashish Vaswani, Noam Shazeer, Niki Parmar, Jakob Uszkoreit, Llion Jones,
  Aidan~N Gomez, {\L}ukasz Kaiser, and Illia Polosukhin. 2017.
\newblock \href
  {https://papers.nips.cc/paper/2017/file/3f5ee243547dee91fbd053c1c4a845aa-Paper.pdf}
  {Attention is all you need}.
\newblock In \emph{Proceedings of NeurIPS}.

\bibitem[{Wang et~al.(2021)Wang, Xing, and Liu}]{wang2021actionclip}
Mengmeng Wang, Jiazheng Xing, and Yong Liu. 2021.
\newblock \href {https://arxiv.org/abs/2109.08472} {{ActionCLIP}: A new
  paradigm for video action recognition}.
\newblock In \emph{Proceedings of CVPR}.

\bibitem[{Wang et~al.(2022{\natexlab{a}})Wang, Yang, Men, Lin, Bai, Li, Ma,
  Zhou, Zhou, and Yang}]{wang2022unifying}
Peng Wang, An~Yang, Rui Men, Junyang Lin, Shuai Bai, Zhikang Li, Jianxin Ma,
  Chang Zhou, Jingren Zhou, and Hongxia Yang. 2022{\natexlab{a}}.
\newblock \href {https://proceedings.mlr.press/v162/wang22al/wang22al.pdf}
  {{OFA}: Unifying architectures, tasks, and modalities through a simple
  sequence-to-sequence learning framework}.
\newblock In \emph{Proceedings of ICML}.

\bibitem[{Wang et~al.(2022{\natexlab{b}})Wang, Yu, Yu, Dai, Tsvetkov, and
  Cao}]{DBLP:journals/corr/abs-2108-10904}
Zirui Wang, Jiahui Yu, Adams~Wei Yu, Zihang Dai, Yulia Tsvetkov, and Yuan Cao.
  2022{\natexlab{b}}.
\newblock \href {https://openreview.net/pdf?id=GUrhfTuf_3} {{SimVLM}: Simple
  visual language model pretraining with weak supervision}.
\newblock In \emph{Proceedings of ICLR}.

\bibitem[{Xie and Zheng(2021)}]{xie2021zsd}
Johnathan Xie and Shuai Zheng. 2021.
\newblock \href {https://arxiv.org/abs/2109.12066} {{ZSD-YOLO}: Zero-shot
  {YOLO} detection using vision-language knowledge distillation}.
\newblock \emph{arXiv preprint arXiv:2109.12066}.

\bibitem[{Xu et~al.(2017)Xu, Zhu, Choy, and Fei-Fei}]{xu2017scene}
Danfei Xu, Yuke Zhu, Christopher~B Choy, and Li~Fei-Fei. 2017.
\newblock \href
  {https://openaccess.thecvf.com/content_cvpr_2017/papers/Xu_Scene_Graph_Generation_CVPR_2017_paper.pdf}
  {Scene graph generation by iterative message passing}.
\newblock In \emph{Proceedings of CVPR}.

\bibitem[{Yang et~al.(2022)Yang, Gan, Wang, Hu, Ahmed, Liu, Lu, and
  Wang}]{yang2021crossing}
Zhengyuan Yang, Zhe Gan, Jianfeng Wang, Xiaowei Hu, Faisal Ahmed, Zicheng Liu,
  Yumao Lu, and Lijuan Wang. 2022.
\newblock \href {https://arxiv.org/abs/2111.12085} {{UniTAB}: Unifying text and
  box outputs for grounded vision-language modeling}.
\newblock In \emph{Proceedings of ECCV}.

\bibitem[{Yao et~al.(2022)Yao, Dong, Zhang, Zhang, Xie, Liu, Lin, Sun, and
  Wang}]{yao2022prompt}
Yuan Yao, Bowen Dong, Ao~Zhang, Zhengyan Zhang, Ruobing Xie, Zhiyuan Liu, Leyu
  Lin, Maosong Sun, and Jianyong Wang. 2022.
\newblock \href {https://aclanthology.org/2022.findings-acl.273.pdf} {Prompt
  tuning for discriminative pre-trained language models}.
\newblock In \emph{Findings of ACL}.

\bibitem[{Yao et~al.(2021{\natexlab{a}})Yao, Zhang, Han, Li, Weber, Liu,
  Wermter, and Sun}]{yao2021visual}
Yuan Yao, Ao~Zhang, Xu~Han, Mengdi Li, Cornelius Weber, Zhiyuan Liu, Stefan
  Wermter, and Maosong Sun. 2021{\natexlab{a}}.
\newblock \href
  {https://openaccess.thecvf.com/content/ICCV2021/papers/Yao_Visual_Distant_Supervision_for_Scene_Graph_Generation_ICCV_2021_paper.pdf}
  {Visual distant supervision for scene graph generation}.
\newblock In \emph{Proceedings of ICCV}.

\bibitem[{Yao et~al.(2021{\natexlab{b}})Yao, Zhang, Zhang, Liu, Chua, and
  Sun}]{yao2021cpt}
Yuan Yao, Ao~Zhang, Zhengyan Zhang, Zhiyuan Liu, Tat-Seng Chua, and Maosong
  Sun. 2021{\natexlab{b}}.
\newblock \href {https://arxiv.org/abs/2109.11797} {{CPT}: Colorful prompt
  tuning for pre-trained vision-language models}.
\newblock \emph{arXiv preprint arXiv:2109.11797}.

\bibitem[{Yu et~al.(2021)Yu, Tang, Yin, Sun, Tian, Wu, and Wang}]{yu2020ernie}
Fei Yu, Jiji Tang, Weichong Yin, Yu~Sun, Hao Tian, Hua Wu, and Haifeng Wang.
  2021.
\newblock \href {https://ojs.aaai.org/index.php/AAAI/article/view/16431/16238}
  {{ERNIE-ViL}: Knowledge enhanced vision-language representations through
  scene graphs}.
\newblock In \emph{Proceedings of AAAI}.

\bibitem[{Yu et~al.(2018{\natexlab{a}})Yu, Lin, Shen, Yang, Lu, Bansal, and
  Berg}]{yu2018mattnet}
Licheng Yu, Zhe Lin, Xiaohui Shen, Jimei Yang, Xin Lu, Mohit Bansal, and
  Tamara~L Berg. 2018{\natexlab{a}}.
\newblock \href
  {https://openaccess.thecvf.com/content_cvpr_2018/papers/Yu_MAttNet_Modular_Attention_CVPR_2018_paper.pdf}
  {{MAttNet}: Modular attention network for referring expression
  comprehension}.
\newblock In \emph{Proceedings of CVPR}.

\bibitem[{Yu et~al.(2016)Yu, Poirson, Yang, Berg, and Berg}]{yu2016modeling}
Licheng Yu, Patrick Poirson, Shan Yang, Alexander~C Berg, and Tamara~L Berg.
  2016.
\newblock \href {https://arxiv.org/abs/1608.00272} {Modeling context in
  referring expressions}.
\newblock In \emph{Proceedings of ECCV}.

\bibitem[{Yu et~al.(2018{\natexlab{b}})Yu, Yu, Xiang, Zhao, Tian, and
  Tao}]{yu2018rethinking}
Zhou Yu, Jun Yu, Chenchao Xiang, Zhou Zhao, Qi~Tian, and Dacheng Tao.
  2018{\natexlab{b}}.
\newblock \href {https://www.ijcai.org/proceedings/2018/0155.pdf} {Rethinking
  diversified and discriminative proposal generation for visual grounding}.
\newblock In \emph{Proceedings of IJCAI}.

\bibitem[{Zellers et~al.(2019)Zellers, Bisk, Farhadi, and
  Choi}]{zellers2019recognition}
Rowan Zellers, Yonatan Bisk, Ali Farhadi, and Yejin Choi. 2019.
\newblock \href
  {https://openaccess.thecvf.com/content_CVPR_2019/papers/Zellers_From_Recognition_to_Cognition_Visual_Commonsense_Reasoning_CVPR_2019_paper.pdf}
  {From recognition to cognition: Visual commonsense reasoning}.
\newblock In \emph{Proceedings of CVPR}.

\bibitem[{Zellers et~al.(2021)Zellers, Lu, Hessel, Yu, Park, Cao, Farhadi, and
  Choi}]{zellers2021merlot}
Rowan Zellers, Ximing Lu, Jack Hessel, Youngjae Yu, Jae~Sung Park, Jize Cao,
  Ali Farhadi, and Yejin Choi. 2021.
\newblock \href
  {https://proceedings.neurips.cc/paper/2021/file/c6d4eb15f1e84a36eff58eca3627c82e-Paper.pdf}
  {{MERLOT}: Multimodal neural script knowledge models}.
\newblock In \emph{Proceedings of NeurIPS}.

\bibitem[{Zellers et~al.(2018)Zellers, Yatskar, Thomson, and
  Choi}]{zellers2018neural}
Rowan Zellers, Mark Yatskar, Sam Thomson, and Yejin Choi. 2018.
\newblock \href
  {https://openaccess.thecvf.com/content_cvpr_2018/papers/Zellers_Neural_Motifs_Scene_CVPR_2018_paper.pdf}
  {Neural motifs: Scene graph parsing with global context}.
\newblock In \emph{Proceedings of CVPR}.

\bibitem[{Zeng et~al.(2022)Zeng, Zhang, and Li}]{zeng2021multi}
Yan Zeng, Xinsong Zhang, and Hang Li. 2022.
\newblock \href {https://arxiv.org/abs/2111.08276} {Multi-grained vision
  language pre-training: Aligning texts with visual concepts}.
\newblock In \emph{Proceedings of ICML}.

\bibitem[{Zhang et~al.(2022)Zhang, Yao, Chen, Ji, Liu, Sun, and
  Chua}]{zhang2022fine}
Ao~Zhang, Yuan Yao, Qianyu Chen, Wei Ji, Zhiyuan Liu, Maosong Sun, and Tat-Seng
  Chua. 2022.
\newblock \href {https://arxiv.org/pdf/2103.15365.pdf} {Fine-grained scene
  graph generation with data transfer}.
\newblock In \emph{Proceedings of ECCV}.

\bibitem[{Zhang et~al.(2021)Zhang, Li, Hu, Yang, Zhang, Wang, Choi, and
  Gao}]{zhang2021vinvl}
Pengchuan Zhang, Xiujun Li, Xiaowei Hu, Jianwei Yang, Lei Zhang, Lijuan Wang,
  Yejin Choi, and Jianfeng Gao. 2021.
\newblock \href
  {https://openaccess.thecvf.com/content/CVPR2021/papers/Zhang_VinVL_Revisiting_Visual_Representations_in_Vision-Language_Models_CVPR_2021_paper.pdf}
  {{VinVL}: Revisiting visual representations in vision-language models}.
\newblock In \emph{Proceedings of CVPR}.

\bibitem[{Zhou et~al.(2022)Zhou, Yang, Loy, and Liu}]{zhou2021learning}
Kaiyang Zhou, Jingkang Yang, Chen~Change Loy, and Ziwei Liu. 2022.
\newblock \href {https://arxiv.org/abs/2109.01134} {Learning to prompt for
  vision-language models}.
\newblock \emph{International Journal of Computer Vision}.

\end{thebibliography}
\bibliographystyle{acl_natbib}

\appendix


\section{Pre-training Details}
\label{sec:appendix}
We provide pre-training details and statistics of the pre-training corpora.

\begin{table*}[!t]
    \centering
    \small
    \setlength{\tabcolsep}{4pt}
    \begin{tabular}[t]{l ccccccc}
    \toprule
     Dataset & RefCOCO & RefCOCO+ & RefCOCOg & Flickr & GQA & VCR & Visual Genome \\
    \midrule
     \# Image-text pairs & 107K & 107K & 72K & 148K & 762K & 1.7M & 1.8M \\
     \# Images & 15K & 15K & 20k & 30K & 63K & 80K & 46K \\
    \bottomrule
    \end{tabular}
    \caption{Statistics of pre-training corpora. Numbers of image-text pairs and images are reported.}
    \label{Table:data_statistics}
\end{table*}

\paragraph{Implementation details.} Our backbone consists of a 6-layer text Transformer encoder, a ViT-B/16 visual encoder, and a 6-layer cross-modal Transformer encoder (commonly referred to as base size in the literature), with 209.5M parameters in total. The backbone is open-sourced for research usage. In pre-training, we initialize PEVL with pre-trained parameters from ALBEF for computation efficiency. PEVL is pre-trained with learning rate 8e-5, batchsize 512 on 32 NVIDIA V100 GPUs for 5 epochs. The number of position tokens is 512, with decay rate $\alpha=0.25$, and weighting hyperparameter $\lambda=2$ in ordering-aware reconstruction. The hyperparameters are selected by grid search on the validation sets. For data augmentation, following MDETR~\cite{kamath2021mdetr}, we augment images with random size crop. We also follow Pix2Seq~\cite{chen2021pix2seq} to adopt horizontal flip to augment images, where ``left'' and ``right'' in text are swapped after flip to ensure the semantic correctness. Previous works suggest that an intermediate in-domain pre-training can better adapt VLP models to downstream tasks~\cite{chen2020uniter}. We therefore conduct an intermediate pre-training before tuning on each downstream task.

\paragraph{Pre-training Corpora.} The pre-training corpora consist of referring expressions~\cite{yu2016modeling,mao2016generation}, Flickr30k~\cite{plummer2015flickr30k}, GQA~\cite{hudson2019gqa}, VCR~\cite{zellers2019recognition} and Visual Genome dense captions~\cite{krishna2017visual}, with 4.7M image-text pairs and 210K images in total. We provide the detailed statistics of the datasets in Table~\ref{Table:data_statistics}.

\section{Downstream Tasks}
We provide details of dataset, prompt tuning and baseline models for each downstream task.
\subsection{Referring Expression Comprehension} 

\paragraph{Datasets.} RefCOCO~\cite{yu2016modeling} is collected from a referential game between two players. The dataset is split into train, validation, testA and testB sets, containing 120,624, 10,834, 5,657 and 5,095 expression-object pairs respectively. RefCOCO+~\cite{yu2016modeling} is also constructed in an interactive fashion, and contains 120,191, 10,758, 5,726 and 4,889 expression-object pairs in train, validation, testA and testB sets respectively. RefCOCOg~\cite{mao2016generation} is built in a non-interactive way, and contains 80,512, 4,896 and 9,602 expression-object pairs in train, validation and test sets respectively.

\paragraph{Prompt Tuning.} We tune the model with learning rate 1e-5, weight decay 0.02, and batchsize 32 for 10 epochs. Following previous works~\cite{dosovitskiy2020image,li2021align}, we use a higher image resolution of 512 in downstream tuning. The hyperparameters are selected by grid search on the validation set for all experiments. During inference, we select the position token with the largest reconstruction score for each of the four masked tokens. 

\paragraph{Baselines.} We compare with state-of-the-art baselines, including MAttNet~\cite{yu2018mattnet},    DDPN~\cite{yu2018rethinking},   VL-T5~\cite{DBLP:conf/icml/ChoLTB21},  ViLBERT~\cite{lu2019vilbert},  UNITER~\cite{chen2020uniter},   VL-BERT~\cite{su2019vl},  VinVL~\cite{zhang2021vinvl},  VILLA~\cite{DBLP:conf/nips/Gan0LZ0020},  ERNIE-ViL~\cite{yu2020ernie},  MDETR~\cite{kamath2021mdetr}, and  ALBEF~\cite{li2021align}. We also compare with two concurrent works that achieve competitive performance on visual grounding tasks, including UniTAB~\cite{yang2021crossing} and OFA~\cite{wang2022unifying}. We adopt accuracy@0.5 as the evaluation metrics, where an expression is considered correctly grounded if the intersection over union between the top prediction and ground truth is greater than 0.5

\subsection{Phrase Grounding}
\paragraph{Datasets.} Flickr30k entities dataset~\cite{plummer2015flickr30k} is collected through annotating 276K entities in the 158K captions from Flickr30k with object bounding boxes. The dataset is split into train, validation and test sets, with 148,915, 14,433, 14,481 noun phrases respectively.
\paragraph{Prompt Tuning. } We tune the model with learning rate 1e-5, weight decay 0.02, and batchsize 128 for 10 epochs. Following previous works \cite{kamath2021mdetr,yang2021crossing}, we evaluate our model under the merged-boxes protocol, where the boxes of a phrase (e.g., \textit{crowd}) referring to multiple objects are merged by their union. We use resolution 512 during downstream tuning. During tuning and inference, our model predicts the bounding box of each object separately.

\paragraph{Baselines. } We compare with state-of-the-art baselines, including DDPN~\cite{yu2018rethinking}, UniTAB~\cite{yang2021crossing} and MDETR~\cite{kamath2021mdetr}. UniTAB~\cite{yang2021crossing} performs multi-task fine-tuning with several downstream task datasets. We adopt accuracy@0.5 as the evaluation metrics.

\smallskip
\subsection{Visual Relation Detection} 
\paragraph{Datasets. } We use Visual Genome~\cite{krishna2017visual} dataset for the evaluation of this task. The dataset is split into train, validation and test sets, with 65,651, 5,000, 32,422 images respectively. The of object categories and relation categories in the dataset are 150 and 50 respectively.

\paragraph{Prompt Tuning. } We tune the model with learning rate 2e-5, weight decay 0.02, and batchsize 256 for 5 epochs. The resolution of images is 512. The ratio of negative samples (i.e., no relations between the object pair) and positive samples is 3:1.

\paragraph{Baselines.} We compare with strong baselines, including MotifNet~\cite{zellers2018neural}, Unbiased~\cite{tang2020unbiased}, GPS-Net~\cite{lin2020gps}, MSDN~\cite{xu2017scene}, VCTree~\cite{tang2019learning}, DT2-ACBS~\cite{desai2021dt2}, VisualDS~\cite{yao2021visual}, and IETrans~\cite{zhang2022fine}. For evaluation metrics, we adopt Recall@K(R@K), which is the ratio of correct relationship in the top K confident relationship predictions, and mean Recall@K(mR@K), which is the average recall upon all predicate classes.

\smallskip
\subsection{Visual Commonsense Reasoning} 
\paragraph{Datasets.} VCR dataset~\cite{zellers2019recognition} is collected through creating questions requiring commonsense reasoning by workers for given images from 110K movie scenes. The dataset is split into train, validation, test sets with 212,923, 26,534, 25,263 questions respectively.
\paragraph{Prompt Tuning. } We tune the model with learning rate 1e-5, weight decay 0.02, and batchsize 4,096 for 5 epochs. We use resolution 512 in downstream tuning. PEVL predicts binary labels indicating the whether candidate is correct, given the text of question concatenated with answer, or question, answer concatenated with rationale. In Q $\rightarrow$ AR, we first predict an answer from four answer candidates, and then pick a rationale from four rationale candidates based on the predicted answer.

\paragraph{Baselines. } We compare with strong baselines, including R2C~\cite{zellers2019recognition}, TAB-VCR~\cite{LinNeurIPS2019} and strong VLP models including VisualBERT~\cite{DBLP:journals/corr/abs-1908-03557}, ViLBERT~\cite{lu2019vilbert}, Unicoder-VL~\cite{li2020unicoder}, VL-BERT~\cite{su2019vl}, B2T2~\cite{alberti2019fusion} and UNITER~\cite{chen2020uniter}. We report the accuracy of predicting the answer (Q → A), rationale (QA → R) and both (Q → AR).

\smallskip
\subsection{Visual Question Answering} 
\paragraph{Datasets.} GQA dataset~\cite{hudson2019gqa} is collected through automatically generating questions and answers with functional programs based on the scene graphs in Visual Genome. The dataset is split into train, validation, test-dev,and test sets, with 14,305,356, 2,011,853, 172,174 and 1,340,048 questions respectively.
\paragraph{Prompt Tuning. } We tune the model with learning rate 1e-5, weight decay 0.02, batchsize 256 for 5 epochs. Following  MDETR~\cite{kamath2021mdetr}, the intermediate pre-training is conducted on the unbalanced train set, and prompt tuning on the balanced train set. The resolution of image is 384. We infer the answers based on the 1,853 candidates from \citet{kamath2021mdetr}.

\paragraph{Baselines.} We compare with existing methods reported on the GQA balanced validation dataset, including LXMERT~\cite{tan2019lxmert}, BAN~\cite{kim2018bilinear}, CTI~\cite{do2019compact} and CFR~\cite{nguyen2021coarse}.

\section{Ethical Considerations}
Potential risks of this work lie in (1) privacy issues of the pre-training images and text from the Web, (2) misuse of the model (e.g., visual relation detection for monitoring human activity), and (3) toxic model outputs. The initial version of this paper is released at \url{https://arxiv.org/abs/2205.11169}. The picture in Figure~\ref{fig:framework} is obtained from the RefCOCO dataset.

\end{document}